\documentclass[letterpaper, 10 pt, conference]{ieeeconf}  

\IEEEoverridecommandlockouts                            

\usepackage{times}
\usepackage{epsfig}
\usepackage{graphicx}
\usepackage{amsmath}
\usepackage{amssymb}
\usepackage{array}
\usepackage{multirow}
\usepackage{epsfig}
\usepackage{booktabs}
\usepackage{tabularx}
\usepackage{color}
\usepackage{subfigure}
\usepackage{bm}
\usepackage{diagbox}
\usepackage[colorlinks=true,linkcolor=blue,citecolor=green,urlcolor=blue,]{hyperref}

\usepackage[ruled,linesnumbered]{algorithm2e}

\title{\LARGE \bf

Stereo-LiDAR Depth Estimation with Deformable Propagation and Learned Disparity-Depth Conversion
}

\author{Ang Li$^1$, Anning Hu$^1$, Wei Xi$^{2,3}$, Wenxian Yu$^1$ and Danping Zou$^{1*}$
\thanks{$^1$ Shanghai Key Laboratory of Navigation and Location-based Service, Shanghai Jiao Tong University. $^2$ Intelligent Perception Institute, Midea Corporate Research Center. $^3$ Blue-Orange Lab, Midea Group. $^*$Corresponding author: Danping Zou ({\tt\small dpzou@sjtu.edu.cn)}. This work was supported by National Key R\&D Program of China (2022YFB3903801) and Midea Group's 3D Vision Project.}
}

\begin{document}

\maketitle
\thispagestyle{empty}
\pagestyle{empty}

\begin{abstract}
Accurate and dense depth estimation with stereo cameras and LiDAR is an important task for automatic driving and robotic perception. While sparse hints from LiDAR points have improved cost aggregation in stereo matching, their effectiveness is limited by the low density and non-uniform distribution. To address this issue, we propose a novel stereo-LiDAR depth estimation network with \underline{S}emi-\underline{D}ense hint \underline{G}uidance, named SDG-Depth. Our network includes a deformable propagation module for generating a semi-dense hint map and a confidence map by propagating sparse hints using a learned deformable window. These maps then guide cost aggregation in stereo matching. To reduce the triangulation error in depth recovery from disparity, especially in distant regions, we introduce a disparity-depth conversion module. Our method is both accurate and efficient. The experimental results on benchmark tests show its superior performance. Our code is available at \href{https://github.com/SJTU-ViSYS/SDG-Depth}{https://github.com/SJTU-ViSYS/SDG-Depth}. 

\end{abstract}

\section{INTRODUCTION}

Dense depth estimation is a fundamental task in autonomous driving, robotic navigation \cite{nalpantidis2010stereo}, and 3D reconstruction \cite{geiger2011stereoscan}. Stereo matching, a widely adopted technique for depth estimation, computes the dense disparity map between two rectified images. The disparity map is then converted into a depth map or a 3D point cloud through triangulation. Learning-based stereo matching methods have achieved impressive performances \cite{zbontar2016stereo} in recent years. 
However, their performances still degrade in the case of severe illumination changes and textureless \cite{shivakumar2019real}. 
Recent studies show that the sparse depth from LiDAR \cite{siddiqui2020extensible}\cite{qiu2019deeplidar}\cite{poggi2019guided} can be used as additional hints to guide stereo matching in challenging scenarios, where the depth hints are taken as additional inputs and processed by the neural network.

\begin{figure}[t]
    \centering
    \subfigure{
    \includegraphics[width=0.96\linewidth]{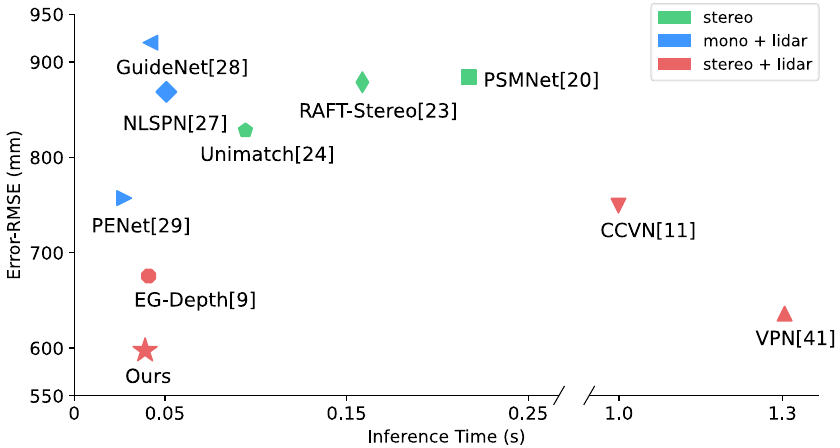}
    }
    \caption{Our network achieves the best trade off in accuracy and inference speed on the KITTI Completion dataset. Green, blue, and red marks represent results from stereo matching, monocular depth completion, and stereo-LiDAR fusion methods, respectively.}
    \label{fig:time_accuracy}
    \end{figure}

However, raw LiDAR depth data are sparse and non-uniformly distributed, making the neural network ignore them or produce noisy predictions. Therefore, the sparse depth usually needs to be expanded and spread to nearby pixels before further processing as demonstrated in prior work such as \cite{huang2021s3}\cite{xu2023expanding}. 
While achieving good results, their methods are based solely on local information, which may not work well in regions with occlusion or object boundaries where the depth is often discontinuous. In those areas, the expanded hints are usually over-smoothing or contain trailing effects, which leads to poor stereo matching results. 

Furthermore, accurately recovering depth at a distance is challenging due to the quadratic growth of triangulation errors with distance. Existing methods \cite{zhang2022slfnet}\cite{xu2023expanding}\cite{wang20193d} primarily focus on enhancing the accuracy of disparity predictions to compensate for triangulation errors. However, further improving disparity predictions typically comes at the cost of more complex networks and increased computational requirements, which may still result in large depth errors via triangulation even when only tiny disparity errors are present for distant points.

To address the aforementioned issues, we propose a novel stereo-LiDAR depth estimation network that adopts a sparse LiDAR deformable propagation module and a learned disparity-depth conversion module, as depicted in Figure \ref{fig:network}. The propagation module computes propagation weights through local self-correlation, which integrates global context and local information. It then propagates this information within learned deformable windows to effectively expand depth hints across occluded and boundary areas. We utilize both the expanded disparity feature and RGB feature to construct a cost volume for stereo matching. Additionally, the expanded disparity guides cost aggregation during stereo matching.
To obtain dense disparity, we perform cost aggregation using a commonly used coarse-to-fine 3D CNN \cite{gu2020cascade}\cite{shen2021cfnet}. To mitigate triangulation errors during disparity-to-depth conversion, we introduce a lightweight network that predicts adjustment residuals in both disparity and depth spaces based on high-frequency features.

We evaluate the proposed network on various benchmark datasets, including KITTI depth completion \cite{uhrig2017sparsity}, Virtual KITTI2 \cite{cabon2020virtual}, and MS2 \cite{shin2023deep} dataset. The experimental results show our method achieves state-of-the-art accuracy and efficiency (see Figure \ref{fig:time_accuracy}).  Our key contributions are highlighted as follows:
\begin{itemize}
\item We introduce a novel stereo-LiDAR fusion network for precise and efficient depth estimation, setting new benchmarks for accuracy and speed across various datasets.

\item We design a sparse LiDAR deformable propagation module to effectively expand depth hints across occlusion and boundary areas. The module propagates sparse hints within learned varying-shaped windows, incorporating global context and local information.

\item We develop a lightweight disparity-depth conversion module that enables precise depth recovery from disparity with low memory and computational requirements.

\end{itemize}

\section{RELATED WORKS}

\subsection{Stereo Matching}

Stereo matching aims to find a dense disparity map between two rectified images, facilitating scene depth recovery through triangulation. Learning-based stereo matching methods have made significant progress. Early approaches \cite{zagoruyko2015learning}\cite{mayer2016large} adopted siamese networks to extract patch-wise features or predict matching costs. Follow-up methods \cite{kendall2017end}\cite{chang2018pyramid} introduced 3D CNN to capture pixel-wise correspondence for more accurate predictions.
To reduce the memory and computational costs of 3D CNN, many methods \cite{xu2020aanet}\cite{shen2021cfnet}\cite{tankovich2021hitnet} have be proposed. Xu et al. \cite{xu2020aanet} used deformable convolution to simplify cost aggregation, and Tankovich et al. \cite{tankovich2021hitnet} utilized coarse-to-fine convolution to reduce computation.

Though existing stereo-matching methods \cite{xu2020aanet}\cite{lipson2021raft}\cite{xu2023unifying} have achieved promising results, they may exhibit decreased performance in challenging scenarios, such as textureless regions, occlusion, and distant objects. Recent research suggests that incorporating LiDAR points can enhance performance, highlighting the potential of using multi-modal information for dense depth estimation.

\subsection{Monocular Depth Completion}
Monocular depth completion aims to predict a dense depth map from a single image and sparse LiDAR points.
Existing methods can be classified into two categories: spatial propagation methods \cite{cheng2018depth}\cite{liu2017learning}\cite{park2020non}\cite{uhrig2017sparsity} and fusion-based methods \cite{tang2020learning}\cite{hu2021penet}\cite{liu2023mff}\cite{teixeira2020aerial}\cite{qiu2019deeplidar}. Spatial propagation methods typically involve learning an affinity matrix based on RGB images to propagate depth values. In contrast, fusion-based methods leverage the geometric information from multi-modal data, including single images and LiDAR points. These methods \cite{tang2020learning}\cite{hu2021penet} typically employed two sub-networks to extract multi-layer features from RGB images and sparse depth, respectively. These features are then fused at different stages to generate depth maps. However, the significant sparsity and non-uniform distribution of LiADR points pose challenges, leading to performance degradation in regions with insufficient LiDAR points.

\subsection{Stereo-LiDAR Fusion}

Stereo-LiDAR fusion methods \cite{zhang2022slfnet}\cite{shivakumar2019dfusenet}\cite{cheng2019noise}\cite{you2019pseudo}\cite{mai2021sparse}  produce more precise depth predictions by combining stereo images and sparse LiDAR points. There are primarily two ways to explore geometric cues from sparse LiDAR points and stereo images: fusing the two-modalities information at the feature level, and leveraging sparse points to guide cost aggregation in stereo matching. Early fusion-based methods \cite{park2018high}\cite{zhang2020listereo} extracted features from sparse depth and stereo images and integrated multi-modal information by concatenating these features. 
Zhang et al. \cite{zhang2022slfnet} constructed an attention map by incorporating image features and depth features for depth prediction.

In contrast to the aforementioned approaches, methods that employ sparse points as guidance explicitly leverage the metric geometric information of LiDAR points. Poggi et al. \cite{poggi2019guided} constructed a Gaussian modulation to regulate the original cost volume. Based on this idea, Huang et al. \cite{huang2021s3} and Xu et al. \cite{xu2023expanding} expanded sparse depth into semi-dense depth as guidance for cost volume. Despite achieving performance improvements, these methods either expand depth within fixed-shape windows or propagate depth based on the original RGB images, resulting in limited performance due to cross-boundary propagation and illumination variations.

Compared to existing stereo methods, we propose a novel and efficient stereo-LiDAR depth estimation network. Specifically, we design a learnable network that propagates sparse LiDAR points within deformable windows, incorporating both global context and local information, to produce semi-dense hints as guidance.  
Moreover, we develop a lightweight disparity-depth conversion module to accurately recover depth from disparity, leveraging high-frequency image information. Experiments show that our method achieves significantly better prediction accuracy and speed than existing methods.

\begin{figure*}[ht]
    \centering
    \subfigure{
    \includegraphics[width=0.81\linewidth]{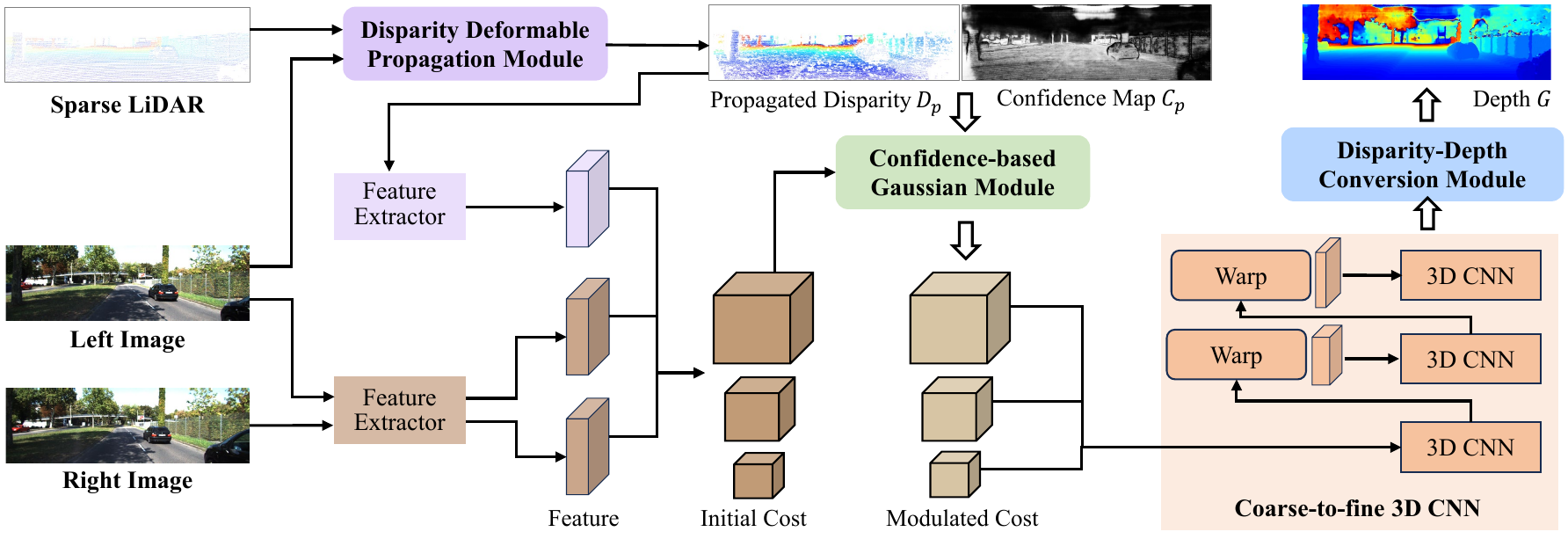}
    }
    \caption{The architecture of our proposed network. Firstly, the disparity \textbf{D}eformable \textbf{P}ropagation (DP) module propagates sparse LiDAR within varying-shaped windows to semi-dense disparity. Based on the generated disparity map and confidence map, the \textbf{C}onfidence-based \textbf{G}aussian (CG) module regulates the cost volume that is constructed from the features of stereo images and expanded disparity. Subsequently, dense disparity is obtained by employing coarse-to-fine 3D CNN on the regulated cost volume. Finally, the learned \textbf{D}isparity-\textbf{D}epth \textbf{C}onversion (DDC) module accurately recovers depth from the disparity of 3D CNN.}
    \label{fig:network}
    \vspace{-10pt}
    \end{figure*}

\section{Method}

Figure \ref{fig:network} shows the overview architecture of our stereo-LiDAR depth estimation network. It mainly consists of four components: 1) a sparse disparity \textbf{D}eformable \textbf{P}ropagation (DP) module, which expands sparse hints to a semi-dense hint map along with its confidence through learned varying-shaped windows. The confidence map indicates the reliability of the propagated hint map; 2) a \textbf{C}onfidence-based \textbf{G}aussian (CG) module, which constructs a Gaussian distribution along the disparity using the expanded semi-dense disparity map and its confidence map to constrain the cost volume effectively; 3) a coarse-to-fine 3D CNN is employed to produce dense disparity from the modulated cost volumes; 4) a \textbf{D}isparity-\textbf{D}epth \textbf{C}onversion (DDC) module, which accurately recovers depth from disparity and reduces triangulation error based on high-frequency features.

\subsection{Deformable Propagation (DP) Module}
\begin{figure}[t]
    \centering
    \subfigure{
    \includegraphics[width=0.9\linewidth]{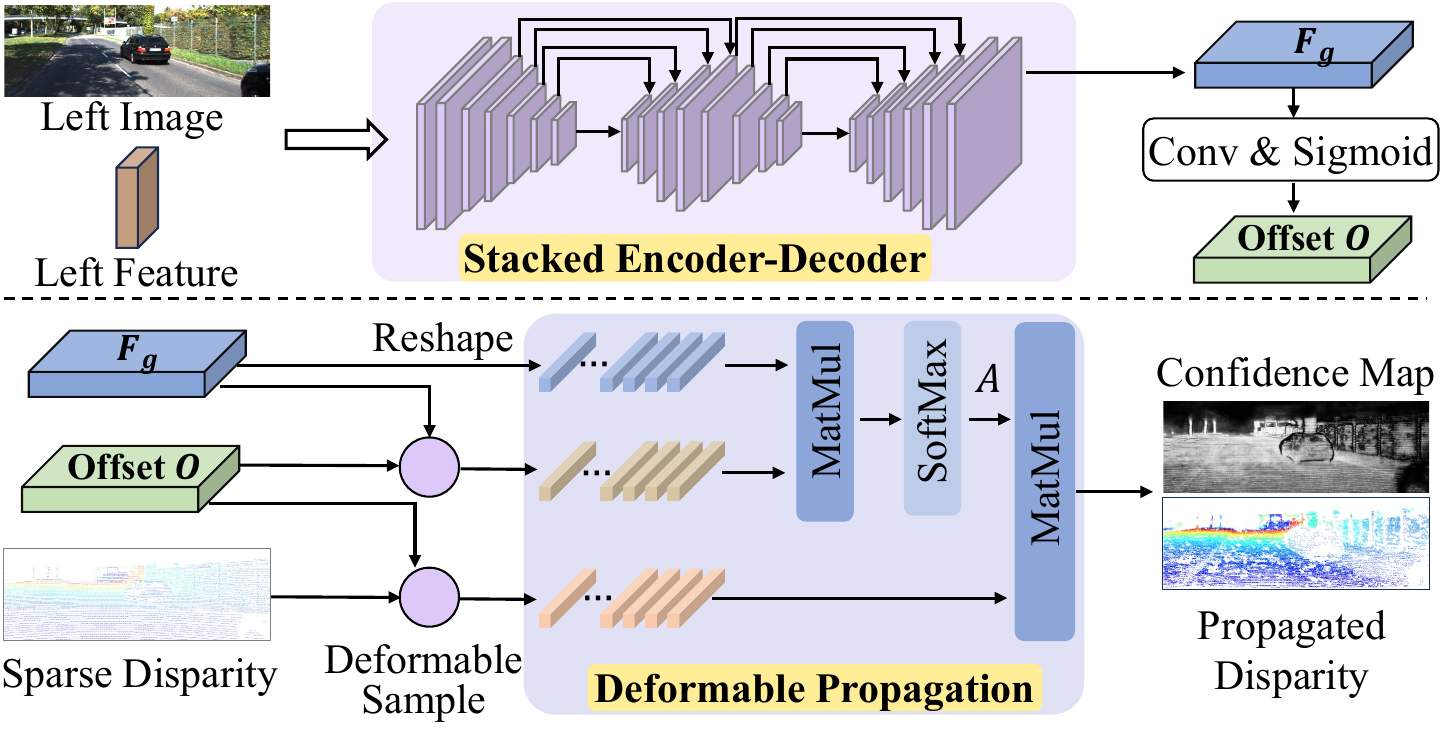}
    }
    \caption{Disparity deformable propagation module. The module computes the propagation weight by employing local self-correlation based on the learned 2D offset field and propagates sparse hints within the deformable windows.}
    \label{fig:apnet}
    \vspace{-10pt}
    \end{figure}

\textbf{Feature extraction with global awareness: }
Given an image $I\in\mathbb{R}^{H \times W \times 3}$ and the corresponding sparse disparity map $D\in\mathbb{R}^{H \times W}$, our objective is to propagate disparity values to the surrounding area based on pixel correlations. We've observed that propagation using features extracted from a limited local region \cite{huang2021s3} often leads to noisy hints.
To address this, as illustrated in Figure \ref{fig:apnet}, we adopt a stacked encoder-decoder architecture to extract features, taking the left image and the left feature used for cost volume construction as inputs. This enables us to encode a larger receptive field of image information into the feature representation $F_g$, facilitating hint propagation with global awareness.

\textbf{Deformable propagation: }Sparse hint propagation within fixed-shape windows usually encounters challenges at object boundaries because depth is usually discontinuous in these regions. Assigning a single depth value across such areas will result in over-smoothness. Inspired by deformable convolution \cite{dai2017deformable}\cite{zhu2019deformable}, we design deformable propagation with local self-correlation \cite{dosovitskiy2020image} to improve propagation across boundaries, as shown in Figure \ref{fig:apnet}. The deformable propagation consists of three steps: 1) generating a learned 2D offset field; 2) computing propagation weights using local self-correlation based on the offset field; and 3) propagating sparse disparity within the deformable windows.

Given the extracted global-aware feature $F_g$, the learned 2D offset field $O$ is produced with a convolution layer and a sigmoid operation.
\begin{equation}
    \label{eq:offset}
      \small   
      O=Sigmoid(Conv(F_g)-0.5)\times 2 \in \mathbb{R}^{H\times W\times P^2 \times 2}
    \end{equation}
where $P$ is size of the propagation window, and $P$ is set to be $9$ in our experiments if not specified. $O$ is rescaled to $[-1,1]$ for stable computation.

The deformable propagation weight $A$ is calculated based on the local self-correlation of the feature $F_g$, which is formulated as follows,
\begin{equation}
    \label{eq:offset}
      \small   
      A=Softmax(F^*_g\psi(F_g,O))\in \mathbb{R}^{H\times W\times P^2}
    \end{equation}
where $F^*_g$ represents the feature reshaped from $F_g$; $\psi$ denotes sampling operation within deformable windows generated based on the offset field.

Finally, the propagated disparity $D_{p}$ and corresponding confidence map $C_{p}$ can be formulated as follows,
\begin{equation}
    \small   
    D_{p}=A\cdot \psi(D,O) \in \mathbb{R}^{H\times W}
\end{equation}
\begin{equation}
    \small   
    C_{p}=A\cdot \psi(M_{sparse},O) \in \mathbb{R}^{H\times W}
\end{equation}
where $M_{sparse} \in \{0,1\}$ represents the valid mask of the sparse disparity $D$. $C_{p}$ indicates the reliability of the propagated depth $D_{p}$.

The deformable propagation is performed at a resolution of $1/4$, ensuring high efficiency.

\subsection{Confidence-based Gaussian Module}

As demonstrated in \cite{poggi2019guided}\cite{huang2021s3}, modulating the cost volume with Gaussian distributions derived from sparse LiDAR points can significantly enhance stereo matching performance. However, the quality of the propagated depth $D_{p}$ varies in different regions, and erroneous propagation may bias stereo matching. Therefore, we extend the Gaussian modulation introduced in \cite{poggi2019guided} to a confidence-based Gaussian modulation, which adapts to the varying reliability of the propagated depth. 

Let the original cost volume constructed from the left feature, semi-dense disparity feature and shift right feature be $CV \in \mathbb{R}^{H\times W \times D_{max} \times L}$, where $D_{max}$ represents the max disparity and $L$ is the number of feature channels; $D_{p} \in \mathbb{R}^{H\times W}$ denotes the propagated semi-dense disparity. Then the Gaussian modulation can be described as:

\begin{equation}
   \small   
CV'(x,y,d)=\emph{f}\cdot CV(x,y,d)
\end{equation}
\begin{equation}
   \small   
   \emph{f}=1-M(x,y)+M(x,y)\cdot k\cdot C_{p}(x,y) \cdot e^{-\frac{(d-D_{p}(x,y))^2}{2\omega^2}}
\end{equation}
where $\emph{f}$ represents the pixel-varying modulation weights; $C_p(x,y)$ describing the confidence at pixel $(x,y)$;  $M \in \{0,1\}$ is a valid mask of $C_p>\rho$ to exclude unreliable propagated depth; $k$ and $\omega$ are constants to adjust the height and width of the Gaussian distribution; $\rho$, $k$ and $\omega$ are set to $0.4$, $2$ and $8$ in our implementation, respectively; $d \in \{0,1,\cdots,D_{max}-1\}$; In this function, when unreliable propagation occurs, $C$ takes on a very small value and the mask $M$ will be $0$, leading the $CV'$ to be the original cost volume, and excluding the erroneous guidance.

\subsection{Coarse-to-fine 3D CNN}
\label{sec:3dcnn}

While Gaussian modulation is applied, the cost volume still exhibits noise that hampers accurate matching. Cost aggregation with 3D CNN is leveraged to incorporate extensive contextual information for precise dense disparity estimation. To alleviate the computational burden, a coarse-to-fine 3D CNN \cite{gu2020cascade}\cite{shen2021cfnet} is employed in our network, where the generated multi-scale disparity maps are used for training losses. Furthermore, we incorporate the searching range adjustment based on disparity uncertainty \cite{shen2021cfnet} into our network to further enhance efficiency. As shown in Figure \ref{fig:time_accuracy}, our network achieves high inference speed through the coarse-to-fine and adaptive searching range strategies.

\subsection{Disparity-Depth Conversion (DDC) Module}

\begin{figure}[t]
    \centering
    \subfigure{
    \includegraphics[width=0.96\linewidth]{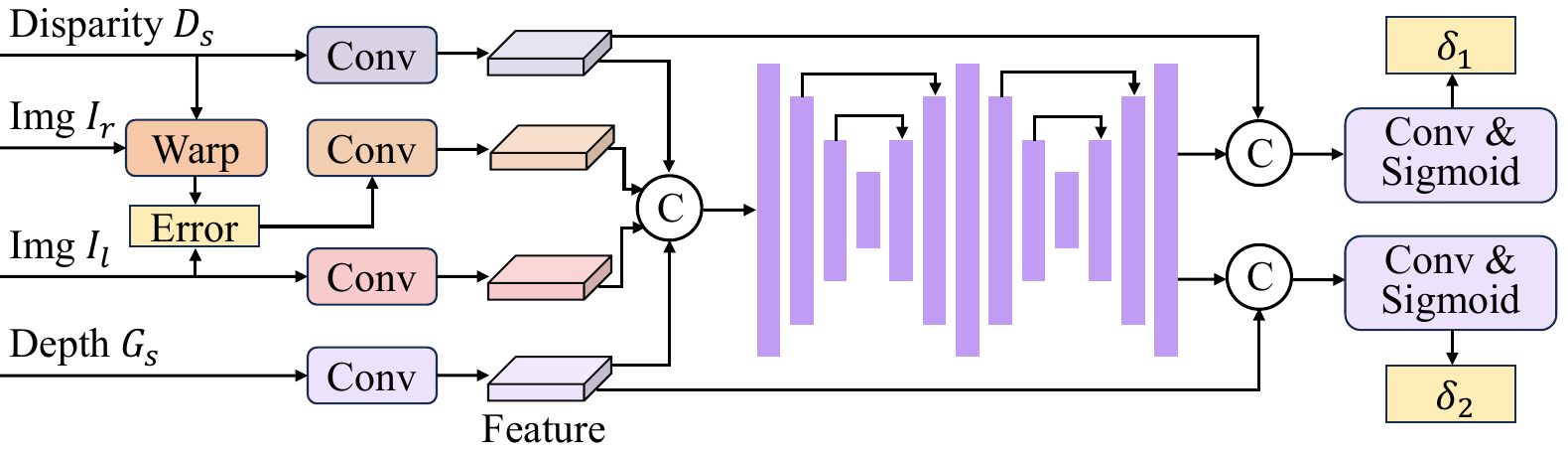}
    }
    \caption{Disparity-depth conversion module. The module generates pixel-wise residuals $\delta_1$ and $\delta_2$ in both the disparity and depth space, based on high-frequency features.}
    \label{fig:ddc}
    \vspace{-15pt}
    \end{figure}

Due to the triangulation error growing quadratically with the distance, recovering scene depth accurately from disparity is challenging, especially in distant regions. To address this challenge, a lightweight disparity-depth conversion module is used for compensating the triangulation error as shown in Figure \ref{fig:ddc}. The network predicts two pixel-wise residuals : $\delta_1$ and $\delta_2$, which are used for compensating disparity and depth errors respectively.
\begin{equation}
    \label{eq:ddc}
    \small   
    G=\frac{b\cdot f}{D_s+\delta_1}+\delta_2 \quad \delta_1, \delta_2\in \mathbb{R}^{H\times W}
 \end{equation}
Here $G \in\mathbb{R}^{H\times W}$  is the converted depth map; $b$ and $f$ are the baseline and focal length of the stereo camera, respectively; $D_s$ is the disparity map from stereo matching.

To enhance edge awareness, we leverage the high-frequency information \cite{zhao2022eai} in the original images to enable the network to acquire error correction capabilities while preserving details. We first warp the right image to the left viewpoint using the disparity $D_s$ obtained from stereo matching, and then compute an error map by comparing the warped right image with the left image. Subsequently, a network with a stacked U-Net-like architecture takes four inputs : the left image, the error map, the disparity $D_s$, and the derived depth $G_s$, and produces the two residuals $\delta_1$ and $\delta_2$ in Equation \ref{eq:ddc}. To ensure stable computation, $\delta_1$ and $\delta_2$ are rescaled to $[-0.2,0.2]$ and $[-0.6,0.6]$, respectively. The entire process can be formulated as follows:

 \begin{equation}
  \small   
  \delta_1, \delta_2=Net(I_l,I_l-Warp(I_r,D_s),D_s,G_s)
\end{equation}

\subsection{Loss Function}
The proposed network is trained end-to-end in a supervised manner. We formulate the loss function by using propagated disparity map $D_p$, disparity maps obtained from different scales of 3D CNN described in Section \ref{sec:3dcnn}, and the depth map $G$ generated from the disparity-depth conversion (DDC) module, as follows,
\begin{equation}
    \small   
    L=a \mathcal{L}_1(D_p,D_{gt})+   b_i\sum_{i=1}^{3}L^i_{disparity}+\lambda L_{depth}(G,G_{gt})
 \end{equation}
 where $\mathcal{L}_1$ represent the L-1 loss function; $L^i_{disparity}$ denotes the L-1 loss based on multi-scale disparity maps of 3D CNN; $L_{depth}$ represents L-1 and L-2 loss of the final depth map $G$. In our experiments, $a$ is set to $0.5$, $b_i, i\in\{1,2,3\}$, are set to $0.5$, $1.0$, and $2.0$, and $\lambda$ is set to $0.7$, respectively.

\begin{figure*}[t]
    \centering
    \subfigure{
    \includegraphics[width=0.97\linewidth]{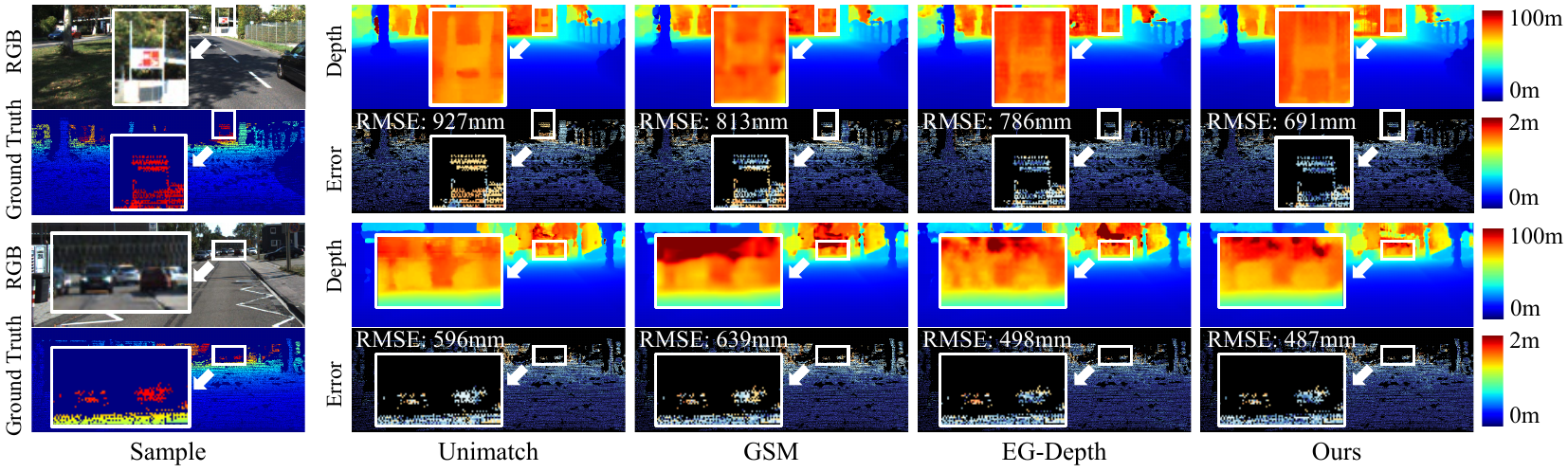}
    }
    \caption{Qualitative results on KITTI depth completion dataset \cite{uhrig2017sparsity}. Our network produces more accurate predictions with smaller depth errors (in blue) and more regular object shapes in distant regions, compared to other state-of-the-art stereo and stereo-LiDAR methods.}
    \label{fig:benchmark_kitti}
    \vspace{-10pt}
    \end{figure*}

\section{EXPERIMENTS}

\subsection{Datasets}

We conduct experiments on three benchmark datasets, including \textbf{KITTI depth completion} \cite{uhrig2017sparsity}, \textbf{Virtual KITTI2} \cite{cabon2020virtual} and \textbf{MS2} \cite{shin2023deep} datasets. KITTI depth completion dataset \cite{uhrig2017sparsity} is a large-scale outdoor driving scenarios dataset. It contains stereo image pairs and raw noise sparse LiDAR points collected by a Velodyne LiDAR. The semi-dense ground-truth depth is obtained by \cite{uhrig2017sparsity}, as shown in Figure \ref{fig:benchmark_kitti}. The dataset provides $42949$ frames for training and $3426$ frames for validation with the image size of $375\times 1242$.

Virtual KITTI2 datasets \cite{cabon2020virtual} is a synthetic dataset and provides dense ground-truth depth maps. The dataset contains five scenes, and we follow \cite{choe2021volumetric} to use "Scene01" and "Scene02" for network training, and the remaining scenes are used for testing. In total, there are $680$ frames for training and $1446$ frames for testing. As in \cite{choe2021volumetric}, we randomly sample points from the dense depth map, resulting in sparse depth maps with a density of $5\%$ which is close to the average density of the sparse depth of KITTI depth completion dataset.

MS2 \cite{shin2023deep} is a large-scale multi-spectral stereo dataset collected in the real world, which provides stereo images, raw LiDAR points, and semi-dense ground-truth depth. Due to the repetitive scenes in the original dataset, we select four splits for training ("2021-08-06-11-23-45", "2021-08-13-16-14-48", "2021-08-13-16-31-10", "2021-08-13-17-06-04") and one split for validation ("2021-08-13-16-08-46"). 
In total, there are $10120$ frames for training and $1272$ frames for validation. The image size is $384\times 1224$.

\subsection{Implementation Details}
We implement our network with PyTroch \cite{paszke2019pytorch} and conduct training on NVIDIA RTX 4090 GPUs. Across all datasets, we train our network using the Adam optimizer with parameters ($\beta_1=0.9,\beta_2=0.999$) and a batch size of $4$. During training, images are cropped to $256\times 512$. For the KITTI completion dataset, the network is trained from scratch for $25$ epochs and the learning rate starts at $10^{-3}$ and reduces to half at epochs $14$, $17$, $19$, and $24$. For Virtual KITTI2 dataset, we fine-tune the network using weights pre-trained on KITTI completion dataset for $5000$ steps, with a learning rate of $3\times 10^{-4}$. For MS2 dataset, we train the network from scratch for $30$ epochs with a constant learning rate of $10^{-3}$. During validation, full-sized original images are fed into the network.

We adopt the standard metrics described in the official KITTI depth completion benchmark \cite{uhrig2017sparsity} to evaluate the quality of the estimated depth maps, including Root Mean Square Error (RMSE), Mean Absolute Error (MAE), and their inverse ones iRMSE and iMAE.

\subsection{Benchmark Evaluation}
\label{sec:benchmark}
We evaluate our network on the three aforementioned benchmarks \cite{uhrig2017sparsity}\cite{cabon2020virtual}\cite{shin2023deep}, comparing it with state-of-the-art stereo, monocular depth completion, and stereo-lidar fusion methods. The quantitative results are presented in Table \ref{tab:kitti completion} for KITTI depth completion dataset, Table \ref{tab:vkitti2} for MS2 and Virtual KITTI2 datasets. The results demonstrate that our method achieves superior performance in most metrics with a higher inference speed as shown in Table \ref{tab:kitti completion}. For example, our method achieves an RMSE of $623.2$, representing a \textbf{21.5\%} reduction compared to the sparse guidance method GSM \cite{poggi2019guided} with an RMSE of $793.4$. Compared to EG-Depth \cite{xu2023expanding} with a similar speed to ours, our method achieves an RMSE of $623.2$, which indicates a \textbf{7.7\%} reduction compared to EG-Depth with an RMSE of $675.5$. The results on Virtual KITTI2 and MS2, as shown in Table \ref{tab:vkitti2},  also demonstrate a significant improvement in the prediction of our method.


\begin{table}[t]    
  \begin{minipage}[t]{0.49\textwidth}
  \caption{Quantitative results on KITTI Depth Completion validate dataset.}   
  \label{tab:kitti completion}
\resizebox{1\textwidth}{!}{ \begin{minipage}{1.3\textwidth}  \small
\centering
\begin{tabular}{lccccccc}
\toprule  
\multicolumn{1}{l}{\multirow{2}{*}{Method}} & \multicolumn{1}{c}{\multirow{2}{*}{Inputs}} & RMSE & MAE & iRMSE & iMAE & FPS & Memory\\
\multicolumn{1}{c}{} & \multicolumn{1}{c}{} & (mm) & (mm) &  (1/km) & (1/km) &(Hz) & (MB)\\
\midrule  
GC-Net \cite{kendall2017end} & S &1031.4 & 405.40&1.681 &1.036 &2.4 &8073 \\
PSMNet \cite{chang2018pyramid} & S & 884.0&332.00 &1.649 & 0.999 &4.6 &4465 \\
RaftStereo\cite{lipson2021raft}& S &878.8 &301.80 &1.751 &0.954 &6.3 &7839 \\
Unimatch \cite{xu2023unifying} &S&828.2 &283.19 &1.666 &0.923 &10.6 &5970  \\
\midrule

GuideNet \cite{tang2020learning} &M+L &920.2 &232.21 &2.318 &0.946 &23.8 &\underline{2139} \\
NLSPN \cite{park2020non} &M+L &868.6 &236.70 &2.614 &1.026&19.7 &\textbf{1939}  \\
PENet \cite{hu2021penet} &M+L &757.2 &209.00 &2.220 &0.920 &\textbf{36.3} &2571 \\
\midrule
Listereo \cite{zhang2020listereo} & S+L &832.2 &283.91 & 2.190&1.100 &- &- \\
GSM \cite{poggi2019guided} & S+L &793.4 &271.48 &1.531 &0.864 &4.4 &4473 \\
CCVN \cite{wang20193d} & S+L &749.3 &252.50 &\textbf{1.397} &0.807 &1.0 &8260 \\
S3 \cite{huang2021s3} & S+L &703.7 &239.60 &1.540 &0.790 &4.3 &6300 \\  
SLFNet \cite{zhang2022slfnet} &S+L &641.1 &\textbf{197.00} &1.773 &0.876 &- &- \\
VPN \cite{choe2021volumetric} & S+L &\underline{636.2} &205.10 &1.872 &0.987 &0.7 &- \\
\hline
\hline
EG-Depth \cite{xu2023expanding} &S+L& 675.5& \underline{197.16}& 1.600&\underline{0.787} &24.4 &5513  \\
\textbf{Ours} &S+L &\textbf{623.2} &197.55 &\underline{1.519} &\textbf{0.772} &\underline{25.6} &5700 \\
\midrule
\end{tabular}

\emph{"S", "M+L" and "S+L" represent stereo camera, monocular camera with LiDAR, and stereo camera with LiDAR, respectively. \textbf{Bold} and \underline{underline} refer to the best and second-best results, respectively.}  
\end{minipage}}
\end{minipage}      
\vspace{-17pt}       
\end{table}

\begin{figure}[t]
  \centering
  \subfigure{
  \includegraphics[width=0.95\linewidth]{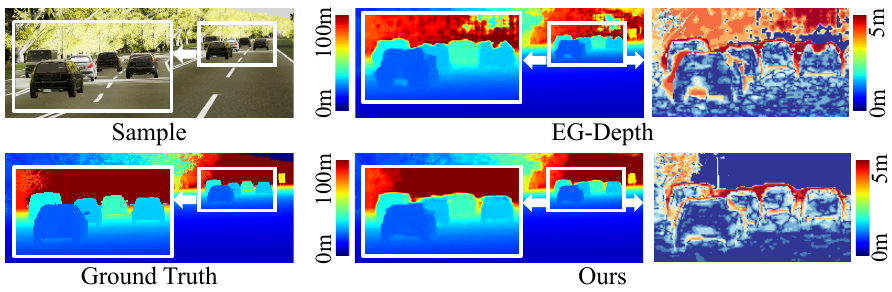}
  }
  \caption{Qualitative results on Virtual KITTI2 dataset \cite{cabon2020virtual}. Our network produces more accurate predictions with sharper edges in distant regions, compared to another state-of-the-art stereo-LiDAR method (EG-Depth) \cite{xu2023expanding}. }
  \label{fig:benchmark_vkitti}
  \vspace{-16pt}
  \end{figure}

To intuitively compare different methods, we show the visual results of different methods on KITTI depth completion dataset in Figure \ref{fig:benchmark_kitti}. It can be observed that our method produces more accurate predictions in distant regions, whereas the other two stereo-lidar methods generate predictions with larger depth errors. Besides, our method generates predictions with a more regular object shape for the cars at a distance (the second sample in Figure \ref{fig:benchmark_kitti}), while other methods failed. Additionally, Figure \ref{fig:benchmark_vkitti} showcases visual results on Virtual KITTI2, highlighting our method's ability to produce depth predictions with sharper edges and more regular object shapes. In summary, our approach generates higher-quality predictions in distant and object edge regions with a lower RMSE. More quantitative and qualitative results are provided in the supplementary material.

\begin{table}[t]
  \begin{minipage}[t]{0.48\textwidth}
      \caption{Quantitative results on MS2 (real-world) \cite{shin2023deep} and Virtual KITTI2 (synthetic) \cite{cabon2020virtual} datasets. 
      }      
      \label{tab:vkitti2}
\centering
\resizebox{0.88\textwidth}{!}{\begin{minipage}{1.1\textwidth}  \small
\begin{tabular}{clcccc}  
  \toprule  
  \multicolumn{1}{c}{\multirow{2}{*}{Dataset}}&\multicolumn{1}{l}{\multirow{2}{*}{Method}} 
  & RMSE & MAE & iRMSE & iMAE \\
  \multicolumn{1}{c}{} & \multicolumn{1}{c}{}
  & (mm) & (mm) &  (1/km) & (1/km) \\
  \midrule  
  
  \multicolumn{1}{c}{\multirow{4}{*}{\shortstack{MS2}}}  
  &Unimatch \cite{xu2023unifying} &1706.72  &946.80  &5.173  &2.519  \\
  &GSM \cite{poggi2019guided} &1434.48 &815.21 &3.494 &1.882  \\        
  &EG-Depth \cite{xu2023expanding} &\underline{1251.01} &\underline{696.00} &\underline{3.284} &\underline{1.763} \\ 
  &Ours  &\textbf{1056.54} &\textbf{604.95} &\textbf{3.209} &\textbf{1.736} \\ 
  \midrule
  \multicolumn{1}{c}{\multirow{6}{*}{\shortstack{Virtual\\KITTI2}}}&Unimatch \cite{xu2023unifying} &3730.59  &1091.4  &7.291  &2.163  \\
  &CCVN \cite{wang20193d}  &3726.83 &915.6 &8.814 &2.456 \\
  &GSM \cite{poggi2019guided} &3510.12 &966.8 &7.059 &1.609 \\
  &VPN \cite{choe2021volumetric}  &3217.16&\underline{712.0}&7.168&2.694\\ 
  &EG-Depth \cite{xu2023expanding} &3184.22 &815.9 &\underline{4.302} &\textbf{1.072} \\   
  &SLFNet \cite{zhang2022slfnet} &\underline{2843.16}&\textbf{696.2}&6.794&2.007\\
  &Ours &\textbf{2821.44} &776.8 &\textbf{4.224} &\underline{1.105} \\ 
  \midrule
  \end{tabular}
\vspace{-5pt}
\end{minipage}}
\end{minipage}
\end{table}

\begin{table}[t]    
  \begin{minipage}[t]{0.48\textwidth}
    \caption{Evaluation on different ranges on KITTI Depth Completion.} 
    \label{tab:range error}
    \centering  
    \resizebox{0.95\linewidth}{!}{\begin{minipage}{1.2\textwidth}\small
    \begin{tabular}{clccccc}
    \hline
    \multirow{2}{*}{ \ \shortstack{Depth Range} \ } & \multirow{2}{*}{Method}      
    & RMSE & MAE & iRMSE  & iMAE  \\ 
    & &(mm)&(mm)&(1/km)&(1/km) \\
    \hline    
    \multicolumn{1}{c}{\multirow{4}{*}{0$-$20m}}
    &Unimatch \cite{xu2023unifying} &268.0  &115.3  &1.706  &0.991  \\  
    &GSM \cite{poggi2019guided}  &237.6 &105.9 &1.643 &0.912  \\     
    & EG-Depth \cite{xu2023expanding} &\underline{228.1} & \underline{96.5}& \underline{1.607} & \underline{0.869}\\
    & Ours &\textbf{227.4} &\textbf{95.8} &\textbf{1.596} &\textbf{0.853}  \\
    
    \hline
    %
    \multirow{4}{*}{20$-$100m}       
    &Unimatch \cite{xu2023unifying} &1711.3  &890.8  &1.458  &0.716  \\ 
    &GSM \cite{poggi2019guided}  &1663.9 &875.2  &1.353 &0.695 \\     
    & EG-Depth \cite{xu2023expanding}  &\underline{1365.3} &\underline{588.5} &\underline{1.200} &\underline{0.498} \\
    & Ours  &\textbf{1284.0} &\textbf{569.9} &\textbf{1.146} & \textbf{0.495}  \\
    \hline  
  
    \end{tabular}
  \end{minipage}}   
    
  \end{minipage}
  \vspace{-12pt}
   
  \end{table}

\subsection{Evaluation on Different Ranges}

We evaluate the performance at different distances to provide a more comprehensive view of the enhancements achieved by our method, as demonstrated in Table \ref{tab:range error}. Our approach exhibits superior accuracy in each region compared to other competitive methods, especially in the distant region. For example, in the distant regions of $20-100m$, our method achieves a $6\%$ reduction in RMSE compared to the second-best method EG-Depth.

\subsection{Ablation}

To verify the effectiveness of each module in our network, we conduct various ablation studies on KITTI depth completion dataset.

\textbf{Ablation on key components: }
To assess the individual contributions of each module, including deformable propagation (DP), confidence-based Gaussian (CG), and disparity-depth conversion (DDC) modules, we remove each module from the entire network and train each model from scratch. The results are shown in Table \ref{tab:ablation_modules}. The removal of any single module leads to a reduction in network performance. By combining all the proposed modules, our network achieves the best performance.
Additionally, it can be observed that although the disparity maps produced by all models are of similar quality (e.g. EPE, D1), the models with the DDC module, such as (a), (b), and (d), predict better depth results compared to (c) without the DDC module. This highlights the substantial contribution of the DDC module, as triangulation introduces errors in the conversion, in contrast, the proposed DDC module enables accurate depth recovery from the disparity obtained from stereo matching, as intended.

\textbf{Ablation on deformable propagation (DP) module: }
The DP module is designed to enhance sparse hints propagation by expanding hints within deformable windows. We conduct ablation experiments on this module, involving propagation within fixed-shape windows and propagation within deformable windows of different sizes. The results are presented in Table \ref{tab:ablation_module_param}. It can be seen that propagation within deformable windows produces superior performance. Besides, the performance of the network decreases when the window size is too large, which is likely due to the network capturing too much contextual information that may be irrelevant to the current pixel, resulting in decreased propagation quality.

\textbf{Ablation on confidence-based Gaussian (CG) module: }We also adjust the hyper-parameter, including height $k$ and width $\omega$ of CG module to achieve optimal performance. As shown in Table \ref{tab:ablation_module_param}, the optimal parameters of $k=2$ and $\omega=8$ are adopted for the best performance.


\begin{table}[t]    
    \begin{minipage}[h]{0.48\textwidth}
  \begin{minipage}[t]{1\textwidth}
    \caption{Ablation on key modules on KITTI depth completion.}
    \label{tab:ablation_modules}
  \centering
\resizebox{0.86\textwidth}{!}{ \begin{minipage}{1.15\textwidth}  \small
\begin{tabular}{@{}llcccccc}
\toprule
\multicolumn{1}{c}{\multirow{3}{*}{}} &
\multicolumn{1}{l}{\multirow{3}{*}{Method}} & \multicolumn{4}{c}{\multirow{1}{*}{Depth map}} &\multicolumn{2}{c}{\multirow{1}{*}{Disparity map}} \\
\cmidrule(r){3-6}
\cmidrule(r){7-8}
\multicolumn{2}{l}{} & RMSE & MAE & iRMSE & iMAE & EPE & D1 \\
\multicolumn{2}{c}{} & (mm) & (mm) &  (1/km) & (1/km) & (pix.) & (\%) \\
\midrule
(a) & w/o DP &688.8 &218.71 &1.589 &0.801 &0.31 &0.15 \\
(b) & w/o CG   &673.2 &219.70 &1.629 &0.815 &0.31 &0.16 \\
(c) & w/o DDC &710.2 &213.66 &1.540 &0.794 &0.30 &0.15  \\
(d) & Ours Full &\textbf{623.2} &\textbf{197.55} &\textbf{1.519} &\textbf{0.772} &\textbf{0.29} &\textbf{0.14} \\
\midrule
\end{tabular}
\end{minipage}}
\end{minipage}
\begin{minipage}[t]{1\textwidth}
  \caption{Ablation study on deformable propagation (DP) and confidence-based Gaussian (CG) modules.}
  \label{tab:ablation_module_param}
    \centering
  \resizebox{0.9\textwidth}{!}{\begin{minipage}{1.1\textwidth}  \small
  \begin{tabular}{clcccc}
      \toprule
      \multicolumn{1}{c}{\multirow{2}{*}{Modules}} &\multicolumn{1}{l}{\multirow{2}{*}{Method}} & RMSE & MAE & iRMSE & iMAE \\
      \multicolumn{1}{c}{} &\multicolumn{1}{c}{}& (mm) & (mm) &  (1/km) & (1/km) \\
      \midrule
      
      \multicolumn{1}{c}{\multirow{5}{*}{\shortstack{\\DP\\Module}}} &        
      w/o offset &649.3 &217.52 &1.563 &0.819  \\
     &with offset &\textbf{623.2} &\textbf{197.55} &\textbf{1.519} &\textbf{0.772}  \\
      \cmidrule(r){2-6}
      &$size=7$ &646.0 &211.64 &1.555 &0.814 \\
      &$size=9$ &\textbf{623.2} &\textbf{197.55} &\textbf{1.519} &\textbf{0.772} \\
      &$size=11$ &635.7 &200.44 &1.537 &0.777 \\
      \midrule
      
      \multicolumn{1}{c}{\multirow{5}{*}{\shortstack{\\CG\\Module}}} &
      w/o confidence &638.9 &206.72 &1.588 &0.811 \\
      &with confidence &\textbf{623.2} &\textbf{197.55} &\textbf{1.519} &\textbf{0.772} \\
      \cmidrule(r){2-6}
      &$k=1,\omega=8$ &644.2 &206.77 &1.556 &0.786 \\
      &$k=2,\omega=8$ &\textbf{623.2} &\textbf{197.55} &\textbf{1.519} &\textbf{0.772} \\ 
      &$k=8,\omega=2$ &642.8 &213.69 &1.573 &0.818 \\
      &$k=8,\omega=1$ &653.7 &214.82 &1.588 &0.819 \\
      
      \midrule
      \end{tabular}
  \vspace{-5pt}
  \end{minipage}}
\end{minipage}
\end{minipage}
\vspace{-5pt}
\end{table}

\section{CONCLUSION}

We present a novel and efficient stereo-lidar depth estimation network. Sparse LiDAR is first adaptively propagated within deformable windows, resulting in a semi-dense disparity map and its corresponding confidence map. Subsequently, to address the variable reliability of propagated disparity, a confidence-based Gaussian module utilizes the semi-dense disparity and confidence map as inputs to guide cost aggregation. Finally, a lightweight module is employed to accurately recover depth from disparity obtained from coarse-to-fine 3D CNN. Comprehensive experiments are conducted in both real-world and synthetic datasets. The results demonstrate the superior performance of our method. Future work includes producing globally consistent scene depth and acquiring real-world datasets with high-precision, high-density ground truth for quantitative evaluation.

\bibliographystyle{IEEEtran}
\bibliography{referencemc}
\end{document}